\documentclass[10pt,twocolumn,letterpaper]{article}

\usepackage{iccv}
\usepackage{times}
\usepackage{epsfig}
\usepackage{graphicx}
\usepackage{amsmath, bm}
\usepackage{amssymb}
\usepackage{subfig}

\usepackage[font=small]{caption}
\usepackage{xspace}
\usepackage{times}
\usepackage{algorithm}
\usepackage{algorithmic}
\usepackage{mathtools}
\usepackage{array}
\usepackage{multirow}
\usepackage[inline]{enumitem}
\usepackage{multirow}
\usepackage{tabularx}
\usepackage{color}
\usepackage[utf8]{inputenc} 
\usepackage[T1]{fontenc}    
\usepackage{url}            
\usepackage{booktabs}       
\usepackage{amsfonts}       
\usepackage{nicefrac}       
\usepackage{microtype}      
\usepackage[dvipsnames]{xcolor}
\usepackage{blindtext}
\usepackage[numbers,sort]{natbib}

\DeclareMathOperator{\softmax}{softmax}
\DeclarePairedDelimiter\norm{\lVert}{\rVert}
\DeclareMathOperator*{\argmax}{arg\,max}

\usepackage[pagebackref=true,breaklinks=true,letterpaper=true,colorlinks,bookmarks=false]{hyperref}

\iccvfinalcopy 

\def\iccvPaperID{6175} 

\ificcvfinal\fi
\begin{document}

\title{Align2Ground: Weakly Supervised Phrase Grounding Guided by \\ Image-Caption Alignment}

\author{
    Samyak Datta$^{\star 1,2}$ \quad
    Karan Sikka$^1$ \quad
    Anirban Roy$^1$ \quad
    Karuna Ahuja$^1$ \\
    Devi Parikh$^{2,3}$ \quad \quad
    Ajay Divakaran$^1$ \quad
    \vspace{3pt}\\
    $^1$SRI International, Princeton, NJ, $^2$Georgia Institute of Technology, $^3$Facebook AI Research\\
    {\tt\small $^2$\{samyak, parikh\}@gatech.edu} \\
    {\tt\small $^1$\{karan.sikka, anirban.roy, karuna.ahuja, ajay.divakaran\}@sri.com}
}

\maketitle

\makeatletter
\DeclareRobustCommand\onedot{\futurelet\@let@token\@onedot}
\def\@onedot{\ifx\@let@token.\else.\null\fi\xspace}

\def\etal{et al\onedot}
\def\etc{etc\onedot}
\def\ie{i.e\onedot}
\def\eg{e.g\onedot}
\def\cf{cf\onedot}
\def\vs{vs\onedot}
\def\pd{\partial}
\def\grad{\nabla}
\def\R{\mathbb{R}}
\def\d{\boldsymbol{\delta}}
\def\y{\textbf{y}}
\def\l{\boldsymbol{\ell}}
\def\wrt{w.r.t\onedot}
\def\a{\boldsymbol{\alpha}}
\def\vertspace{0.6em}
\newcommand{\mat}[1]{\bm{#1}}

\def\mF{\mat{F}}
\def\mM{\mat{M}}
\def\mW{\mat{W}}
\def\mI{\mat{I}}
\def\conv{\circledast}
\def\wsmm{WS2MM}
\def\vx{\mat{x}}
\def\vc{\mat{c}}
\def\vrc{\mat{r_c}}
\def\vp{\mat{p}}
\def\mWl{\mat{W_l}}
\def\sC{{\mathbb{C}}}
\def\sI{{\mathbb{I}}}

\definecolor{redcol}{rgb}{1, 0, 0}
\definecolor{bluecol}{rgb}{0, 0, 1}
\newcommand{\red}[1]{\textcolor{redcol}{#1}}
\newcommand{\blue}[1]{\textcolor{bluecol}{#1}}
\renewcommand{\paragraph}[1]{\smallskip\noindent{\bf{#1}}}

\def\algorithmautorefname{Algorithm}
\def\figureautorefname{Figure}
\def\tableautorefname{Table}
\def\equationautorefname{Eq.}
\def\sectionautorefname{Section}
\def\subsectionautorefname{Section}
\def\subsubsectionautorefname{Section}

\parskip=3pt
\abovedisplayskip 3.0pt plus2pt minus2pt%
\belowdisplayskip \abovedisplayskip
\renewcommand{\baselinestretch}{0.98}

\newenvironment{packed_enum}{
\begin{enumerate}
  \setlength{\itemsep}{1pt}
  \setlength{\parskip}{0pt}
  \setlength{\parsep}{0pt}
}
{\end{enumerate}}

\newenvironment{packed_item}{
\begin{itemize}
  \setlength{\itemsep}{1pt}
  \setlength{\parskip}{0pt}
  \setlength{\parsep}{0pt}
}{\end{itemize}}

\newlength{\abstractReduceTop}
\newlength{\abstractReduceBot}

\newlength{\sectionReduceTop}
\newlength{\sectionReduceBot}

\newlength{\subsectionReduceTop}
\newlength{\subsectionReduceBot}

\newlength{\subsubsectionReduceTop}
\newlength{\subsubsectionReduceBot}

\newlength{\captionReduceTop}
\newlength{\captionReduceBot}

\newlength{\eqnReduceTop}
\newlength{\eqnReduceBot}

\newlength{\horSkip}
\newlength{\verSkip}

\newlength{\figureHeight}
\setlength{\figureHeight}{1.7in}

\setlength{\horSkip}{-.09in}
\setlength{\verSkip}{-.1in}

\setlength{\abstractReduceTop}{-0.16in}
\setlength{\abstractReduceBot}{-0.16in}

\setlength{\sectionReduceTop}{-0.09in}
\setlength{\sectionReduceBot}{-0.08in}

\setlength{\subsectionReduceTop}{-0.10in}
\setlength{\subsectionReduceBot}{-0.12in}

\setlength{\subsubsectionReduceTop}{-0.08in}
\setlength{\subsubsectionReduceBot}{-0.10in}

\setlength{\eqnReduceTop}{-0.05in}
\setlength{\eqnReduceBot}{-0.05in}

\setlength{\captionReduceTop}{-0.13in}
\setlength{\captionReduceBot}{-0.15in}

\vspace{\abstractReduceTop}
\begin{abstract}

We address the problem of grounding free-form textual phrases by using weak supervision from image-caption pairs.  We
	propose a novel end-to-end model that uses caption-to-image retrieval as a ``downstream'' task to guide the
	process of phrase localization.  Our method, as a first step, infers the latent correspondences between
	regions-of-interest (RoIs) and phrases
	in the caption and creates a discriminative image representation using these matched RoIs. In the
	subsequent step, this learned representation is aligned with the caption.  Our key contribution lies in
	building this ``caption-conditioned'' image encoding which tightly couples both the tasks and allows the weak
	supervision to effectively guide visual grounding. We provide extensive empirical and qualitative
	analysis to investigate the different components of our proposed model and compare it with competitive
	baselines. For phrase localization, we report improvements of $4.9\%$ and $1.3\%$ (absolute) over prior state-of-the-art
	on the VisualGenome and Flickr30k Entities datasets. We also report results that are at par with the state-of-the-art on the
	downstream caption-to-image retrieval task on COCO and Flickr30k datasets.

\end{abstract}
\vspace{\abstractReduceBot}

\vspace{\sectionReduceTop}
\section{Introduction}
\label{subsec-intro}

\renewcommand*{\thefootnote}{$\star$}
\setcounter{footnote}{1}
\footnotetext{A major part of this work was done by S. Datta when he was an intern at SRI International, Princeton, NJ}
\renewcommand*{\thefootnote}{\arabic{footnote}}
\setcounter{footnote}{0}

We focus on the problem of visual grounding which involves connecting natural language descriptions with image regions.
Supervised learning approaches for this task entail significant manual efforts in collecting annotations for region-phrase correspondence
~\cite{mao2016generation, wang2018learning}. Therefore, in this work, we address the problem of grounding
free-form textual phrases under weak supervision from only image-caption
pairs~\cite{xiao2017weakly, zhu2016visual7w, rohrbach2016grounding, karpathy2015deep}.

A key requirement in such a weakly supervised paradigm is a tight coupling
between the task for which supervision is available (image-caption matching) and
the task for which we do not have explicit labels (region-phrase matching).
This joint reasoning ensures that the supervised loss from the former is able to effectively guide the
learning of the latter.

Recent works~\cite{karpathy2015deep, karpathy2014deep} have shown evidence that operating under
such a paradigm helps boost performance for image-caption matching. Generally, these models consist
of two stages: (1) a local matching module that infers the latent region-phrase correspondences to generate
local matching information, and (2) a global matching module that uses this information to
perform image-caption matching. This setup allows phrase grounding to act
as an intermediate and a prerequisite task for image-caption matching.
It is important to note that the primary objective of such works has been
on image-caption matching and \emph{not} phrase grounding.

\renewcommand*{\thefootnote}{$\star \star$}
\setcounter{footnote}{2}
\footnotetext{``Young girl holding a kitten'' by Gennadiy Kolodkin is licensed under CC BY-NC-ND 2.0.}
\renewcommand*{\thefootnote}{\arabic{footnote}}
\setcounter{footnote}{0}

\begin{figure}[t]
	\centering
	\includegraphics[width=\columnwidth]{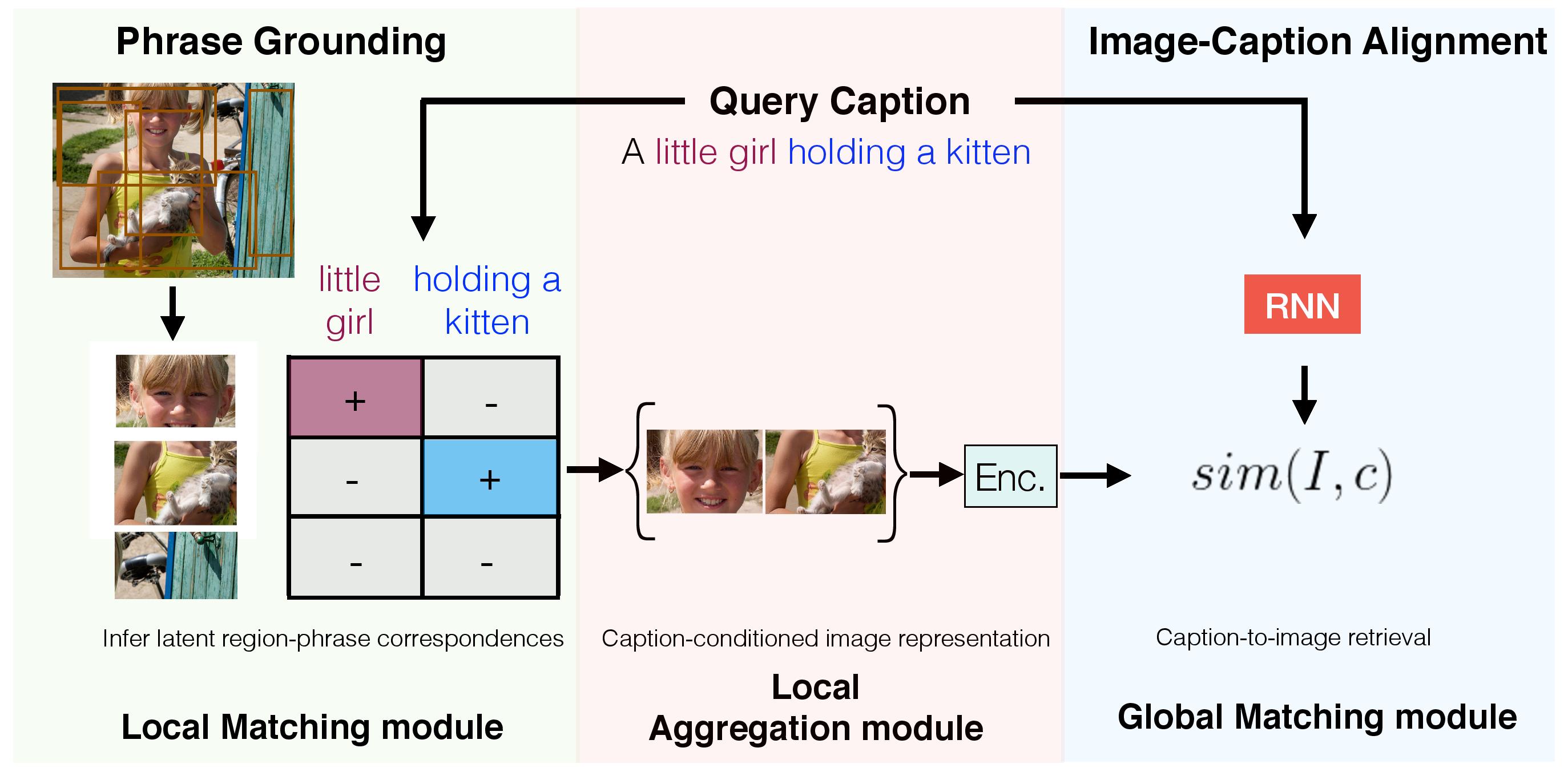}
	\caption{This figure$^{\star \star}$ shows a high-level overview of the proposed Align2Ground model
	which learns to ground phrases by using weak supervision from
	image-caption pairs. It first matches the phrases with local
	image region, aggregates these matched RoIs to generate a
	\emph{caption-conditioned} image representation. It uses this encoding
	to perform image--caption matching.}
	\label{fig:teaser}
	\vspace{-1.5em}
\end{figure}

An artifact of training under such a paradigm is the amplification of correlations between selective regions and
phrases.  For example, a strong match for even a small subset of phrases in the first stage would translate
to a high overall matching score for the image and the entire caption in the second stage.
As a consequence, the model is able to get away with not
learning to accurately ground \emph{all} phrases in an image.
Hence, such a strategy is not an effective solution if the primary aim is visual grounding.
Such ``cheating'' tendencies, wherein models learn to do
well at the downstream task without necessarily getting better at the intermediate task, has also been seen in prior
works such as~\cite{doersch2015unsupervised, kottur2017natural, misra2016shuffle}.

We argue that this ``selective amplification'' behavior is a result of how the local matching information from the first
stage is transferred to the second stage -- via average pooling of the RoI--phrase matching scores.
We address this limitation by proposing a novel mechanism to relay this information about the latent, inferred
correspondences in a manner that enables a much tighter coupling between the two stages.
Our primary contribution is the introduction of a \emph{Local Aggregation Module} that takes the subset of
region proposals that match with phrases and encodes them to get a \emph{caption-conditioned} image representation
that is then used directly by the second stage for image-caption matching (\autoref{fig:teaser}).  We encode the matched proposal features using a permutation-invariant set encoder to get the image
representation. Our novelty lies in designing this effective transfer of information between the supervised and unsupervised
parts of the model such that the quality of image representations for the supervised matching task is a direct
consequence of the correct localization of all phrases.

\begin{figure}[t]
	\centering
	\includegraphics[width=\columnwidth]{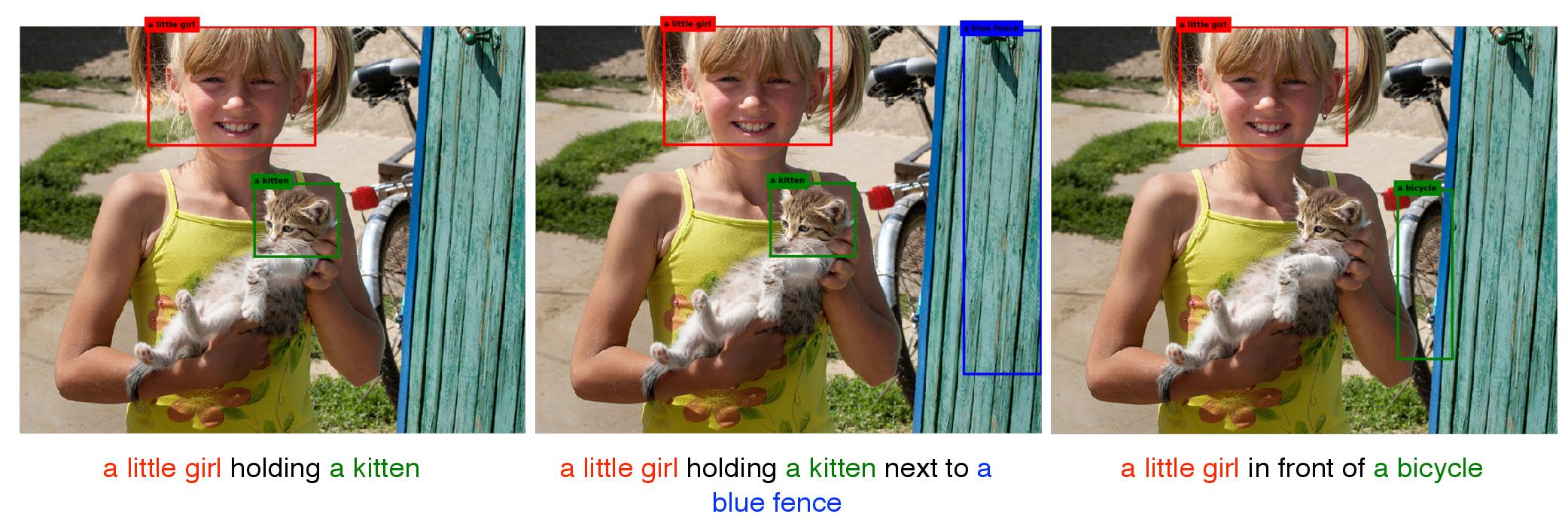}
	\caption{
  For a given image, we show the regions that match with phrases
  from three different query captions, as predicted by our model.
  Our proposed Local Aggregator module computes a caption-conditioned image
  representation by encoding the features of only the matched image regions.
  It is evident that in order for this representation to do well at
  image-caption matching, the grounding of caption-phrases should be proper.
  }
  \label{fig:cap-cond-repr}
  \vspace{-1.5em}
\end{figure}

Our empirical results indicate that such an enforcement of the proper grounding of all phrases via caption-conditioned image
representations (\autoref{fig:cap-cond-repr}) does indeed lead to a better phrase localization performance
(\autoref{tab:loc-sota}, \ref{tab:loc-sota-flickr}).
Moreover, we also show that the proposed discriminative representation allows us to achieve
results that are comparable to the state-of-the-art on the downstream image-caption matching task on both COCO and
Flickr30k datasets (\autoref{tab:c2i-sota}).
This demonstrates that the proposed caption-conditioned representation not only serves as a mechanism
for the supervised loss to be an effective learning signal for phrase grounding, but also does not compromise on the
downstream task performance.

The contributions of our paper are summarized as follows.
\vspace{-.22in}
\begin{packed_item}
	\item
		We propose a novel method to do phrase grounding by using weak supervision from the image-caption
		matching task.
		Specifically, we design a novel \emph{Local Aggregation Module} that computes a \emph{caption-conditioned} image
		representation, thus allowing a tight coupling between both the tasks.

	\item We achieve state-of-the-art performance for phrase
	localization. Our model reports absolute improvements of
	$4.9\%$ and $1.3\%$ over prior state-of-the-art
	on Visual Genome and Flickr30k Entities respectively.

	\item We also report state-of-the-art results on the (downstream) task of
	caption-to-image retrieval on the Flickr30k dataset and obtain performance which
	is comparable to the state-of-the-art on COCO.
\end{packed_item}

\vspace{\sectionReduceTop}
\section{Related Work}

\textbf{Visual-Semantic Embeddings}~\cite{kiros2014unifying, faghri2017vse++, frome2013devise, vendrov2015order,
wang2018learning, ahuja2018understanding, bansal2018zero} have been successfully applied to multimodal tasks such as image-caption retrieval. These methods
embed an (entire) image and a caption in a shared semantic space, and employ triplet-ranking loss based objectives to
fine-tune the metric space for image-caption matching. \cite{faghri2017vse++} further improves this learned, joint
embedding space by using techniques such as hard negative sampling, data augmentation, and fine-tuning of the visual
features.  In addition to these joint embedding spaces, \cite{wang2018learning}  also proposes a similarity network to
directly fuse and compute a similarity score for an image-caption pair.  In contrast to our proposed model, none of
these approaches reason about local structures in the multimodal inputs \ie words/phrases in sentences and regions in
images.

The \textbf{Phrase Localization} task involves learning the correspondences between text phrases and image regions from
a given training set of region--phrase mappings~\cite{mao2016generation, plummer2015flickr30k}. A major challenge
in these tasks is the requirement of ground-truth annotations which are expensive to collect and prone to human error.
Thus, a specific focus of recent work~\cite{rohrbach2016grounding, xiao2017weakly, chen2018knowledge} for phrase localization has been on
learning with limited or no supervision.
For example, \cite{rohrbach2016grounding} learns to leverage the bidirectional correspondence between regions and phrases by
reconstructing the phrases from the predicted region proposals. However, their learning signal is only guided by
the reconstruction loss in the text domain. ~\cite{chen2018knowledge} improves upon their work by adding consistency in both the visual and the
text domains, while also adding external knowledge in the form of distribution of object labels predicted from a
pretrained CNN. As opposed to using hand-annotated phrases (as in the above methods), our model directly makes use of the readily available,
aligned image-caption pairs for visual grounding.

Some prior works also use supervision from image-caption training pairs to perform phrase localization
~\cite{karpathy2014deep, karpathy2015deep, engilberge2018finding, xiao2017weakly, cirik2018using}.
They either rely on using image--caption matching as a downstream task \cite{engilberge2018finding,
karpathy2015deep, karpathy2014deep} or use the sentence parse-tree structure to guide phrase localization.
\cite{engilberge2018finding} achieves phrase localization by first learning a joint embedding space, and then generating
and aggregating top-k feature maps from the visual encoders (conditioned on the text encoding) to find the best matching
spatial region for a query phrase. ~\cite{xiao2017weakly, cirik2018using} propose to use the parse tree
structure of sentences to provide additional cues to guide the model for phrase localization. Among this family of
approaches, our proposed model is conceptually most similar to Karpathy \etal \cite{karpathy2015deep}.
These methods aggregate local region-phrase alignment scores to compute a global
image-caption matching score.
As noted in \autoref{subsec-intro}, such a strategy is able to correctly match an image-caption pair without actually learning to
ground all phrases inside the caption leading to a suboptimal visual grounding.
Our proposed model tackles this issue by
by directly using the matched RoIs to build a discriminative image representation which is able to tightly couple the phrase localization task with the supervised downstream task.

Our work is also closely related to~\cite{niu2017hierarchical} where they not only map images and captions,
but also phrases and regions in the same dense visual-semantic embeddings space. In contrast, our model
provides a clean disentanglement between the region-phrase and the image-caption matching tasks, where
the first stage of our model localizes the phrases and the second stage matches images and captions during training.

\vspace{\sectionReduceBot}

\section{Approach}
\label{sec:app}

\begin{figure*}[t]
	\centering
	\includegraphics[width=\textwidth, scale=0.4]{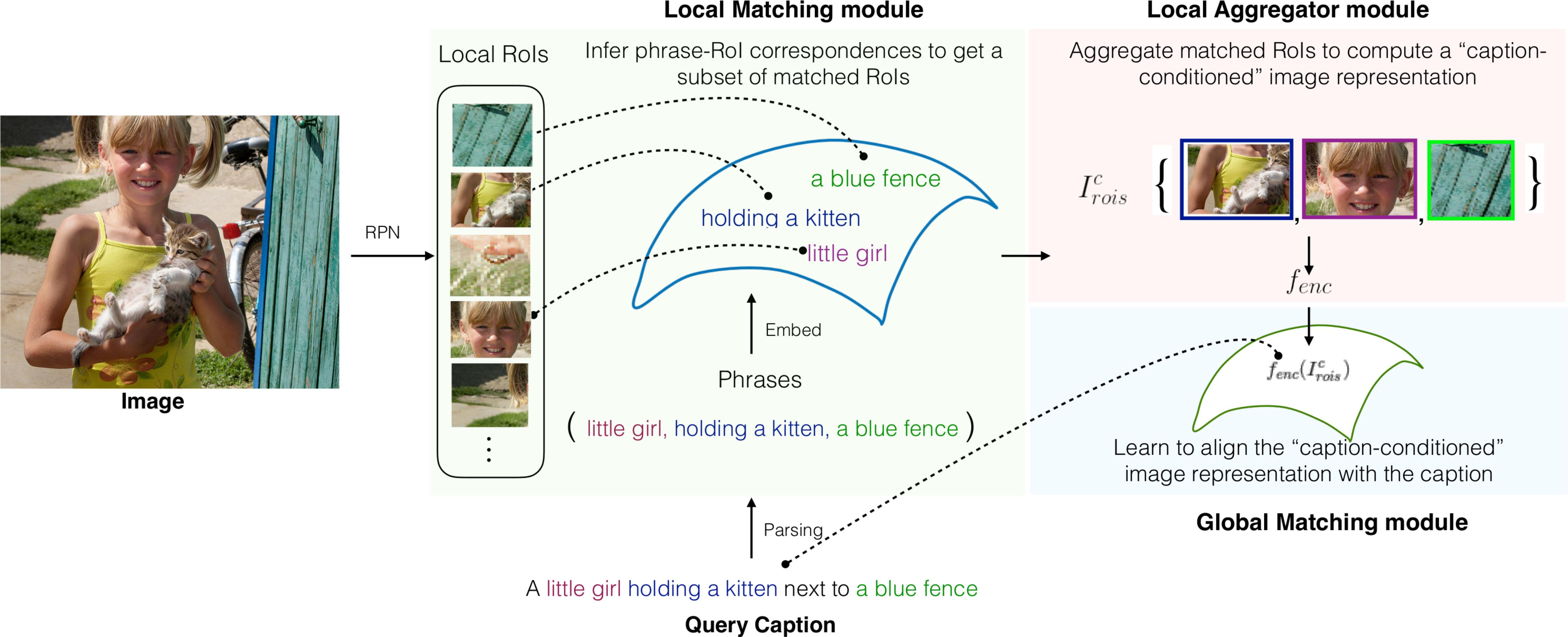}
	\caption{\small{This figure gives a detailed overview of our proposed architecture. The outputs from a Region Proposal Network (RoIs) and the shallow parser (phrases) are fed into the Local Matching module which infers the latent phrase-RoI correspondences. The Local Aggregator module then digests these matched RoIs to create a discriminative, caption-conditioned visual representation -- which is then used to align the image-caption pairs in the Global Matching module.  }}
	\label{fig:flow}
\end{figure*}

We work in a setting where we are provided with a dataset of image-caption pairs for training. We denote an image and a
caption from a paired sample as $I$ and $c$ respectively. For each image, we extract a set of $R$ region proposals, also
referred to as Regions-of-Interest (RoIs), using a pre-trained object detector. We use a pre-trained deep CNN to compute
features for these RoIs and denote them as $\{\vx_j\}_{j=1}^{R}$, where $\vx_j \in \R^{d_v}$. We perform a shallow
parsing (or chunking) of the caption by using an off-the-shelf parser~\cite{collobert2011natural} to obtain $P$ phrases.
We encode each phrase using a Recurrent Neural Network (RNN) based encoder, denoted as $\Phi_{RNN}$. We denote the
encoded phrases as a sequence $(\vp_k)_{k=1}^{P}$, where $\vp_k \in \R^{d_s}$ and $k$ is a positional index for the
phrase in the caption.  Note that we operate in a weakly supervised regime \ie during training, we do not assume
ground-truth correspondences between image regions and phrases in the caption.

During inference, our primary aim is to perform visual grounding. Given a query phrase $p$ and an image $I$, the learned
model identifies the RoI that best matches with the query. We now describe our proposed approach along with the
loss function and the training procedure.

\subsection{Align2Ground}
\label{subsec-approach-wsmm}

We follow the general idea behind prior works that learn to match images and captions by inferring latent alignments
between image regions and words/phrases in the caption~\cite{karpathy2014deep,karpathy2015deep}. These methods operate
under the assumption that optimizing for the downstream task of ranking images with respect to captions requires learning to
accurately infer the latent alignment of phrases with regions \ie phrase grounding.  Specifically, these methods~
\cite{karpathy2014deep,karpathy2015deep} match image-caption pairs by first associating words in the caption to relevant
image regions based on a scoring function.  Thereafter, they average these local matching scores to compute the
image-caption similarity score, which is used to optimize the loss. We shall refer to these methods as Pooling-based
approaches due to their use of the average pooling operation. As discussed earlier, averaging can result in a model that performs
well on the image–caption matching task without actually learning to accurately ground \textit{all} phrases.

In contrast to such methods, our proposed model uses a novel technique that builds a discriminative image representation
from the matched RoIs and uses this representation for the image-caption matching. Specifically, the image
representation that is used to match an image with a caption is conditioned \textit{only on the
subset} of image regions that align semantically with all the phrases in that caption. We argue that such an
architectural design primes the supervision available from image-caption pairs to be a stronger learning signal for visual grounding as
compared to the standard Pooling-based methods. This is a consequence of the explicit aggregation of matched RoIs in our
model which strongly couples both the local and global tasks leading to better phrase localization.

Conceptually, our model relies on three components (see \autoref{fig:flow}) to perform the phrase grounding
and the matching tasks: (1) The Local
Matching Module infers the latent RoI--phrase correspondences for all the phrases in the query caption, (2) The Local
Aggregator Module takes the matched RoIs (as per the alignments inferred by the previous module) and computes a
\emph{caption-conditioned} representation for the image, and (3) The Global Matching Module takes the caption-conditioned
representation of the image and learns to align it with the caption using a ranking loss.  We now describe these modules
in detail.

\noindent \textbf{The Local Matching module} is responsible for inferring the latent correspondences between regions in an
image and phrases from a caption. We first embed both RoIs and phrases in a joint embedding space to measure their
semantic similarity. To do so, we project the RoI $\vx_j$ in the same space as the phrase embeddings, $\vp_{k}$, via a
linear projection. We then measure the semantic similarity, $s_{jk}$, between region $\vx_{j}$ and phrase $\vp_{k}$
using cosine similarity.
\begin{align}
\hat{\vx_{j}} &= \mWl^T \vx_{j} \\
s_{jk} &= \frac{\hat{\vx_{j}}^T \vp_k}{\norm{\hat{\vx_j}}_2\norm{\vp_k}_2}
\end{align}

where $\mWl \in \R^{d_v \times d_s}$ is the projection matrix.

A straightforward approach to infer the matched RoI for a phrase is to the select the top scoring box
\ie for a phrase $p_k$ the matched RoI is $\vx_{j^*}$, where  $j^{*} = \argmax_j s_{jk}$. However, it has
been shown that such a strategy is prone to overfitting since the model often keeps on choosing the same erroneous
boxes~\cite{bilen2014weakly}.
We also take inspiration
from the recent advances in neural attention, and compute attention
weights $\alpha_{jk}$ for each RoI based on a given phrase. We then generate an attended region vector as a linear combination
of the RoI embeddings, weighted by the attention weights.
\begin{align}
	\alpha_{jk} = \softmax(s_{jk})_{j=1}^{R} \quad \vx_k^c = \sum_j \alpha_{jk} \vx_j
\end{align}
Despite the success of this strategy for other multimodal tasks, we found that it is not an effective solution for the
given problem. This is because the discriminativeness of the matched RoI seems to get compromised by the weighted
averaging of multiple matched RoIs during training. We instead propose to add diversity to the training procedure by first
selecting top-k ($k=3$) scoring RoI candidates and then randomly selecting one of them as the matched RoI for the query
phrase. We observe that this strategy adds more robustness to our model by allowing it to explore diverse options during
training.

This module returns a list of caption-conditioned RoIs
$I_{rois}^{c}=(\vx_k^c)_{k=1}^{P}$, where $\vx_k^c$ is the feature vector for the aligned RoI for phrase $p_k$ in
caption $c$.

\paragraph{The Local Aggregator module} uses the RoIs matched in the previous step to generate a caption-conditioned
representation of the image. In contrast to Pooling-based methods, we explicitly aggregate these RoIs to build a more
discriminative encoding for the image.  This idea takes inspiration from adaptive distance metric learning based
approaches~\cite{ye2007adaptive} where the learned distance metric (and equivalently, the embedding space) for computing
similarities is conditioned on the input query. In our case, the image representation is conditioned on the
query caption that we are trying to measure its similarity with.

We propose to use an order-invariant deep encoder to aggregate the RoIs~\cite{zaheer2017deep, qi2017pointnet}. Our
choice is motivated by the assumption of modeling a caption as an orderless collection of phrases.
Such an assumption is justified because a match between a set of phrases and image regions should be
invariant to the order in which those phrases appear in the caption. These different orders might be
generated by say, swapping two noun phrases that are separated by a conjunction such as ``and''.
We implement this encoder, denoted as $f_{enc}$, by using a two-layer Multilayer Perceptron (MLP) with a \emph{mean}
operation~\cite{zaheer2017deep}.  During experiments, we also compare our model with a order-dependent encoding by using
a GRU encoder.  The caption-conditioned image representation, which encodes of the set of matched RoIs, is then passed onto
the next module.
The primary contribution of this work is this module that build this caption-conditioned image
representation and thus ensures a strong coupling between the (unsupervised)
RoI-phrase matching and the supervised image-caption matching task.

\paragraph{The Global Matching module} uses the caption-conditioned image encoding obtained from the Local Aggregator module
and aligns it with the query caption.
We measure similarity between the proposed image representation and the query caption by first embeddings the caption $c$, encoded
by $\Phi_{RNN}$, in the same output space as the image representation by using a two-layer MLP.
We then compute cosine similarity between the two multimodal representations.
\begin{gather}
	\hat{\vc} = MLP(\Phi_{RNN}(c)) \quad \hat{\vrc} = f_{enc}(I^{c}_{rois}) \\
	S_{Ic} = \frac{\hat{\vc}^T \hat{\vrc}}{\norm{\hat{\vc}}_2\norm{\hat{\vrc}}_2}
\end{gather}
\vspace{\eqnReduceBot}

$S_{IC}$ is the similarity between image $I$ and caption $c$.

\paragraph{Loss Function:} We train our model with max-margin ranking loss that enforces the score between
a caption and a paired image to be higher than a non-paired image and vice-versa. Similar to Faghri
\etal~\cite{faghri2017vse++}, we sample the hardest negatives in the mini-batch while generating triplets for the
ranking loss.
\begin{gather}
	\mathcal{L} = \max_{c^{\prime} \notin \sC_I}(0, m - S_{Ic}+ S_{Ic^{\prime}} ) + \max_{I^{\prime} \notin
	\sI_c} (0, m - S_{Ic} + S_{I^{\prime}c})
\end{gather}
where $m$ is the margin, $\sC_I$ is the set of captions paired with image $I$, and $\sI_c$ is the set of images paired
with caption c.
\vspace{\sectionReduceBot}

\section{Experiments}
\label{sec-exp}

In this section, we discuss the experimental setup used to evaluate our model. We first outline the datasets and the
evaluation metrics used to measure performance. We then provide implementation details for our method and a couple
of relevant prior works that our model is conceptually related to.
Next, we establish the benefits of our model by reporting quantitative results
for the phrase localization and the caption-to-image retrieval tasks. We follow that with qualitative results to
provide useful insight into the workings of our model. Finally, we compare our model with several
state-of-the-art methods on both the tasks of phrase localization and caption-to-image retrieval

\subsection{Dataset and Evaluation Metrics}
\label{subsec-dataset}

\begin{table*}[t]
	\centering
  \setlength\tabcolsep{2pt}
  \renewcommand{\arraystretch}{0.72}
  \resizebox{0.85\linewidth}{!}{
  \begin{tabular}{l l l l c c c c l c l l c c c c c l c}
    & & & & \multicolumn{6}{c}{\textbf{\scriptsize{COCO}}} &&& \multicolumn{6}{c}{\textbf{\scriptsize{Flickr30k}}} \\
    \cmidrule{5-10}\cmidrule{13-18}
    & & & & \multicolumn{4}{c}{\scriptsize{Caption-to-Image retrieval}} && \scriptsize{Phrase} &&&
          \multicolumn{4}{c}{\scriptsize{Caption-to-Image retrieval}} && \scriptsize{Phrase} \\
    & & & & \scriptsize$R@1$ & \scriptsize$R@5$ & \scriptsize$R@10$ & \scriptsize Med r &&
          \scriptsize$Det. \%$ &&&
    \scriptsize$R@1$ & \scriptsize$R@5$ & \scriptsize$R@10$ & \scriptsize Med r && \scriptsize$Det. \%$ \\
    \toprule
    \multicolumn{3}{c}{\scriptsize{Global}} & & \scriptsize{$39.3$} & \scriptsize{$74.8$} & \scriptsize{$86.3$} &
          \scriptsize{$2$} && \scriptsize{$12.2$} &&& \scriptsize{$27.1$} & \scriptsize{$56.0$} &
          \scriptsize{$68.4$} & \scriptsize{$4$} && \scriptsize{$8.0$} \\ \multicolumn{3}{c}{\scriptsize{Pooling-based (words)}} & & \scriptsize{$47.9$} & \scriptsize{$81.7$} & \scriptsize{$91.0$} & \scriptsize{$2$}  && \scriptsize{$10.7$} &&& \scriptsize{$40.7$} &
          \scriptsize{$71.2$} & \scriptsize{$80.9$} & \scriptsize{$2$} && \scriptsize{$8.4$} \\
    \multicolumn{3}{c}{\scriptsize{Pooling-based (phrases)}} & & \scriptsize{$48.4$} & \scriptsize{$81.7$} &
          \scriptsize{$91.2$} & \scriptsize{$2$} && \scriptsize{$10.8$} &&& \scriptsize{$41.4$} &
          \scriptsize{$71.4$} & \scriptsize{$81.2$} & \scriptsize{$2$} && \scriptsize{$8.9$} \\

    \cmidrule{1-18}
	  \multicolumn{18}{c}{\scriptsize{\textbf{Align2Ground}}} \\
    \cmidrule{1-18}

    \multirow{6}{*}{\rotatebox[origin=c]{90}{\scriptsize{Proposed model}} $\begin{dcases} \\ \\ \\ \\
    \end{dcases}$} & \multirow{3}{*}{\scriptsize{permInv}} & \scriptsize{max} & & \scriptsize{$40.3$} & \scriptsize{$76.3$}
          & \scriptsize{$87.8$} & \scriptsize{$2$} && \scriptsize{$14.5$} &&& \scriptsize{$29.1$} &
          \scriptsize{$60.8$} & \scriptsize{$72.7$} & \scriptsize{$3$} && \scriptsize{$11.5$} \\
    & & \scriptsize{topk} & & \scriptsize{$56.6$} & \scriptsize{$84.9$} & \scriptsize{$92.8$} &
          \scriptsize{1} && \scriptsize{$14.7$} &&& \scriptsize{$49.7$} & \scriptsize{$74.8$} & \scriptsize{$83.3$} &
          \scriptsize{$2$} && \scriptsize{$11.2$} \\
    & & \scriptsize{attention} & & \scriptsize{$42.8$} & \scriptsize{$78.1$} & \scriptsize{$89.1$} &
          \scriptsize{$2$} && \scriptsize{$10.2$} &&& \scriptsize{$37.9$} & \scriptsize{$67.0$} & \scriptsize{$77.8$} &
      \scriptsize{$2$} && \scriptsize{$6.2$} \\
    \cmidrule{5-18}
    & \multirow{3}{*}{\scriptsize{sequence}} & \scriptsize{max} & & \scriptsize{$39.4$} & \scriptsize{$75.0$} &
          \scriptsize{$87.1$} & \scriptsize{$2$} && \scriptsize{$14.5$} &&& \scriptsize{$29.9$} &
          \scriptsize{$60.9$} & \scriptsize{$72.7$} & \scriptsize{$3$} && \scriptsize{$11.5$} \\
          & & \scriptsize{topk} & & \scriptsize{$58.4$} & \scriptsize{$86.1$} & \scriptsize{$93.5$} &
          \scriptsize{1} && \scriptsize{$14.5$} &&& \scriptsize{$47.9$} & \scriptsize{$75.6$} & \scriptsize{$83.5$} &
          \scriptsize{$2$} && \scriptsize{$11.3$} \\
    & & \scriptsize{attention} & & \scriptsize{$41.9$} & \scriptsize{$77.1$} & \scriptsize{$88.4$} & \scriptsize{$2$} && \scriptsize{$9.8$} &&& \scriptsize{$38.2$} & \scriptsize{$68.4$} & \scriptsize{$78.2$} &
          \scriptsize{$2$} && \scriptsize{$5.6$} \\

    \bottomrule

  \end{tabular}}\\[3pt]
  \caption{Phrase localization and Caption-to-Image retrieval results for models trained on COCO and Flickr30k datasets.
  Note that we report phrase localization numbers on VisualGenome in all the cases.  We compare our proposed model
  (\emph{permInv-topk}) with two prior methods and with different choices for the Local Matching module
  (max/topk/attention) and the Local Aggregator module (permInv/sequence) as discussed in \autoref{sec:app}.}
  \label{tab:results}
  \vspace{-1.5em}
\end{table*}

\paragraph{COCO}~\cite{lin2014microsoft} dataset consists of $123,287$ images with $5$ captions
per image. The dataset is split into $82,783$ training, $5,000$ validation and $5,000$ test images. Following
recent works~\cite{faghri2017vse++, karpathy2015deep, niu2017hierarchical}, we use the standard splits~
\cite{karpathy2015deep} and augment the training set with $30,504$ images from the validation set, that were not
included in the original $5,000$-image validation split.

\paragraph{Flick30k}~\cite{plummer2015flickr30k,young2014image} dataset consists of $31,783$ images with $5$ captions
per image. Additionally, the Flickr30k Entities dataset contains over 275$k$
bounding boxes corresponding to phrases from the captions.
Following prior work, we also use $1,000$ images each for validation and test set, and use the remaining
images for training.

\paragraph{VisualGenome (VG)}~\cite{krishna2017visual} dataset is used to evaluate phrase localization. We use a
subset of images from VG that have bounding box annotations for textual phrases. This subset contains images that
are present in both the validation set of COCO and VisualGenome and consists of $17,359$ images with
$860,021$ phrases.

\paragraph{Metrics:} Since the design of our model uses weak supervision from image--caption pairs to perform phrase
localization, we evaluate our model on two tasks-- (1) phrase localization, and (2) caption-to-image retrieval (C2I).

For \textbf{phrase localization}, we report the percentage of phrases that are correctly localized with respect to the
ground-truth bounding box across all images, where correct localization means $IoU \geq 0.5$~\cite{plummer2015flickr30k}.  We refer to this metric
as phrase localization/detection accuracy ($Det.\%$).  Prior works on visual grounding have also demonstrated localization using
attention based heat maps~\cite{engilberge2018finding, xiao2017weakly}.
As such, they use a \emph{pointing game} based evaluation metric, proposed in~\cite{zhang2018top}, which defines a hit if the
center of the visual attention map lies anywhere inside the ground-truth box and reports the percentage accuracy of
these hits. We compare our model with these prior works by reporting the same which we refer to as the $PointIt \%$ metric
on the Visual Genome and Flickr30k Entities datasets.

For the \textbf{C2I} task, we report results using standard metrics-- (i) Recall@K ($R@K$) for K = 1, 5 and 10 that measures the
percentage of captions for which the ground truth image is among the top-K results retrieved by the model, and (ii)
median rank of the ground truth image in the ranked list of images retrieved by the model.  For C2I retrieval
experiments, we train and evaluate our models using both COCO and Flickr30k datasets.

\subsection{Implementation Details}
\label{subsec-implementation}

\paragraph{Visual features:} We extract region proposals for an image by using Faster-RCNN
~\cite{ren2015faster} trained on both objects and attributes from VG, as provided by Andreson
\etal\footnote{\url{https://github.com/peteanderson80/bottom-up-attention}}~\cite{anderson2018bottom}.
For every image, we select the top $30$ RoIs based on Faster-RCNN's class detection score
(after non-maximal suppresion and thresholding). \footnote{We also experimented with other region proposal methods
such as EdgeBoxes and Faster-RCNN trained on COCO, but found this to be much better.}
We then use RoIAlign~\cite{he2017mask} to extract features ($d_v = 2048$-d) for each of these RoIs using a
ResNet-152 model pre-trained on ImageNet~\cite{he2016identity}.

\paragraph{Text features:} We perform shallow parsing (also known as chunking) using the SENNA
parser~\cite{collobert2011natural} to parse a caption into its constituent phrases. Shallow parsing of sentences first
identifies the constituent parts (such as nouns, verbs) of a sentence and then combines them into higher-order
structures (such as noun-phrases and verb-phrases). In our current work, a phrase generally
comprises noun(s)/verb(s) with modifiers such as adjective(s) and/or preposition(s).
Additionally, we perform post-processing steps based on some handcrafted heuristics (refer to supplementary for more
details) to get the final set of phrases from the captions.

Both phrases and sentences are encoded by using a $2$-layer, bidirectional GRU~\cite{chung2015gated} with a hidden
layer of size $1024$ and using inputs from a $300$ dimensional word embeddings.
We train the word-embeddings from scratch to allow for a fair comparison
with prior work~\cite{collobert2011natural,kiros2014unifying}. We also
experimented with a variant that uses pre-trained GloVe embeddings and found
that the performance is worse than the former.

\paragraph{Prior Works and Align2Ground:} We compare our model with two competing works.  The first method,
refered to as \emph{Global}, embeds both the image and caption in a joint embedding space and computes their matching
score using cosine similarity \cite{faghri2017vse++}.  We also implement the Pooling-based
method~\cite{karpathy2015deep}, that computes similarity between image--caption pairs by summarizing the local
region-phrase matching scores. We use our Local Matching module to infer phrase--region correspondences and then average
these scores. Following the original implementation by Karpathy \etal~\cite{karpathy2015deep}, we first encode the
entire caption using a GRU and then compute the embeddings for each word by using the hidden state at the corresponding
word index (within that caption). We refer to this approach as \emph{Pooling-based (words)}. We also implement a variant
that uses phrases, as used in our method, instead of words (\emph{Pooling-based (phrases)}). For a fair comparison we
use the same image and text encoders for the baselines as well as our model.

\begin{table}[t]
  \centering
  \setlength\tabcolsep{1pt}
  \begin{tabular}{l l  c l c l c l l l c l c l c}
    && \multicolumn{5}{c}{\textbf{COCO}} &&&& \multicolumn{5}{c}{\textbf{Flickr30k}} \\
    \cmidrule{3-7}\cmidrule{11-15}
    && $R@$ && $R@5$ && $R@10$ &&&& $R@1$ && $R@5$ && $R@10$ \\
    \toprule
    DVSA~\cite{karpathy2015deep} && $27.4$ && $60.2$ && $74.8$ &&&& $15.2$ && $37.7$ && $50.5$ \\
    UVS~\cite{kiros2014unifying}&& $31.0$ && $66.7$ && $79.9$ &&&& $22.0$ && $47.9$ && $59.3$ \\
    m-RNN~\cite{mao2014deep} && $29.0$ && $42.2$ && $77.0$ &&&& $22.8$ && $50.7$ && $63.1$ \\
    m-CNN~\cite{ma2015multimodal} && $32.6$ && $68.6$ && $82.8$ &&&& $26.2$ && $56.3$ && $69.6$ \\
    HM-LSTM~\cite{niu2017hierarchical} && $36.1$ && -- && $86.7$ &&&& $27.7$ && -- && $68.8$ \\
    Order~\cite{vendrov2015order} && $37.9$ && -- && $85.9$ &&&& -- && -- && -- \\
    EmbeddingNet~\cite{wang2018learning} && $39.8$ && $75.3$ && $86.6$ &&&& $29.2$ && $59.6$ && $71.7$ \\
    sm-LSTM~\cite{huang2017instance} && $40.7$ && $75.8$ && $87.4$ &&&& $30.2$ && $60.4$ && $72.3$ \\
    Beans~\cite{engilberge2018finding}&& $55.9$ && $86.9$ && $94.0$ &&&& $34.9$ && $62.4$ && $73.5$ \\
    2WayNet~\cite{eisenschtat2017linking} && $39.7$ && $63.3$ && -- &&&& $36.0$ && $55.6$ && -- \\
    DAN~\cite{nam2016dual} && -- && -- && -- &&&& $39.4$ && $69.2$ && $79.1$ \\
    VSE++~\cite{faghri2017vse++} && $52.0$ && -- && $92$ &&&& $39.6$ && -- && $79.5$ \\
    SCAN~\cite{lee2018stacked} && $\mathbf{58.8}$ && $\mathbf{88.4}$ && $\mathbf{94.8}$ &&&& $48.6$ && $\mathbf{77.7}$
			       && $\mathbf{85.2}$ \\
    \cmidrule{1-15}
	  \textbf{Ours} && $56.6$ && $84.9$ && $92.8$ &&&& $\mathbf{49.7}$ && $74.8$ && $83.3$ \\
    \bottomrule
  \end{tabular}\\[3pt]
  \caption{Comparison with the state-of-the-art on the downstream caption-to-image retrieval task.}
  \label{tab:c2i-sota}
  \vspace{-1.5em}
\end{table}

To highlight the effectiveness of using the proposed \emph{topk} scheme for the Local Matching module, we compare it
against both \emph{attention} and \emph{max} based methods as discussed in \autoref{sec:app}.  We also compare the
orderless pooling scheme proposed for the Local Aggregator module with an order-dependent pooling scheme based on a
bidirectional GRU ($2$-layer, hidden layer of size $256$ units). For the orderless pooling scheme we use a $2$-layer
MLP with a hidden layer of size $256$.

We train all models for $60$ epochs with a batch size of $32$ using the Adam optimizer and a learning rate of $0.0002$.
We use a margin of $0.1$ for the triplet-ranking loss in all our experiments. We select the final checkpoints on
the basis of the model's best performance on a small validation set for both localization and C2I tasks.
We warm-start our model by initializing it with a model variant that is in spirit similar to the Pooling-based
methods (during our experiments, we observed that it otherwise takes a long time for it to converge).

\begin{table}[t]
  \caption{Phrase Localization on Visual Genome}
  \label{tab:loc-sota}
  \vspace{-1.5em}
  \renewcommand{\arraystretch}{0.5}
  \centering
  \setlength\tabcolsep{2pt}
	\resizebox{0.85\columnwidth}{!}{%
    \begin{tabular}{c c c c c c c c c}
      \multicolumn{9}{c}{} \\
      \toprule
      Random && Center && Linguistic && Beans In && \multirow{2}{*}{\textbf{Ours}} \\
      (baseline) && (baseline) && Structure~\cite{xiao2017weakly} && Burgers~\cite{engilberge2018finding} && \\
      \midrule
      {$17.1$} && {$19.5$} && {$24.4$} && {$33.8$} && $\textbf{38.7}$ \\
      \bottomrule
    \end{tabular}%
  }
  \vspace{-1.5em}
\end{table}

\subsection{Quantitative Results}

We now report the quantitative performance of prior methods as well as different design choices within the proposed
model in~\autoref{tab:results}. We start by comparing Pooling-based methods with the Global method. We then discuss the
impact of using the proposed matching and aggregating strategies against other choices in our model on the phrase
localization and the C2I tasks. We report all results in absolute percentage points.

\begin{figure*}
	\setlength\tabcolsep{.03in}
	\begin{tabular}{lllll}

    \subfloat[\small \textcolor{cyan}{a fork} next to \textcolor{red}{an apple}, \textcolor{GreenYellow}{orange} and \textcolor{BurntOrange}{onion}]{\includegraphics[width = 0.8in, height=1.25in]{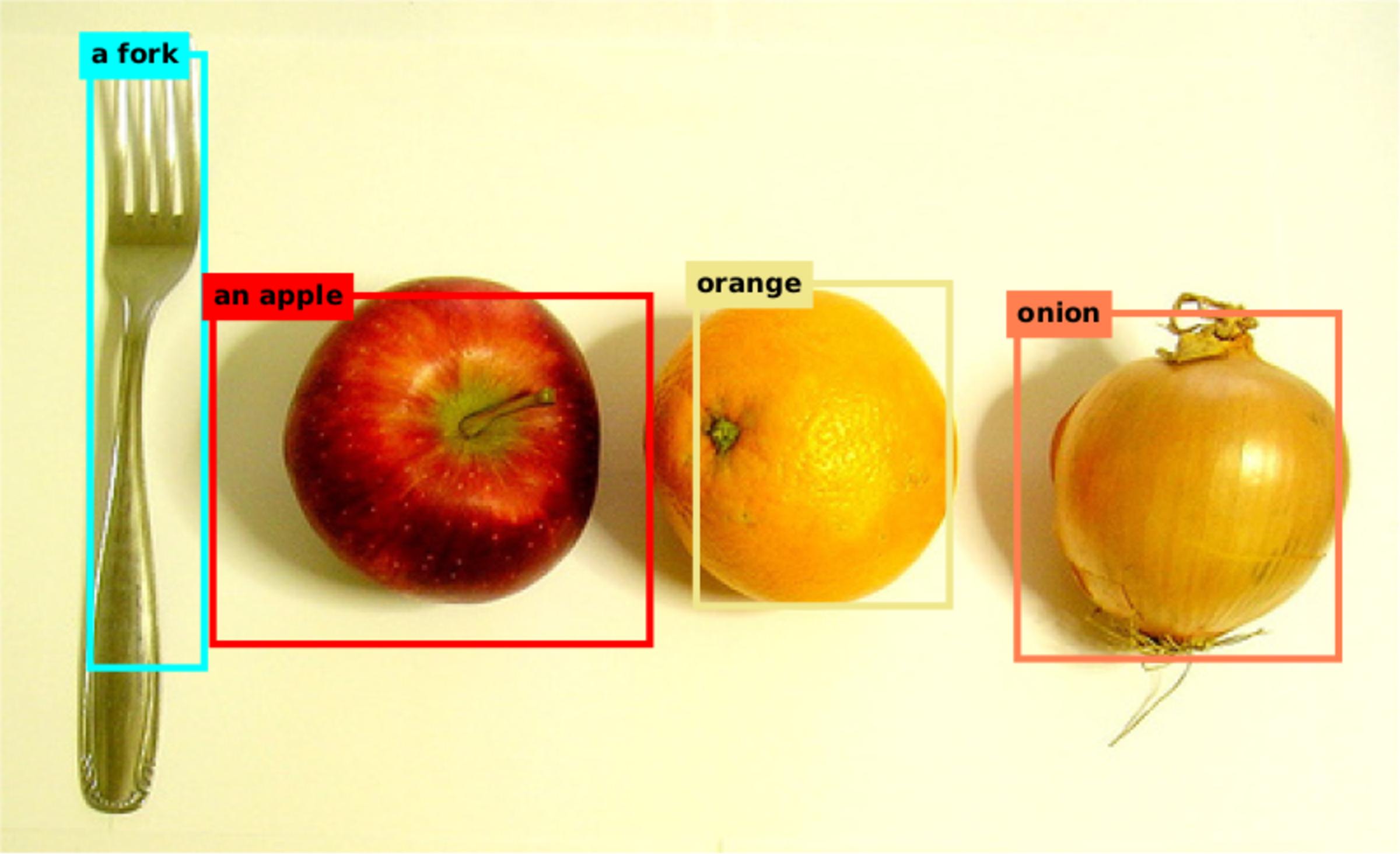}}
		&
		\subfloat[\small \textcolor{Maroon}{cat drinking water} from \textcolor{red}{a sink} in \textcolor{OliveGreen}{a bathroom}
		]{\includegraphics[width = 1in, height=1.25in]{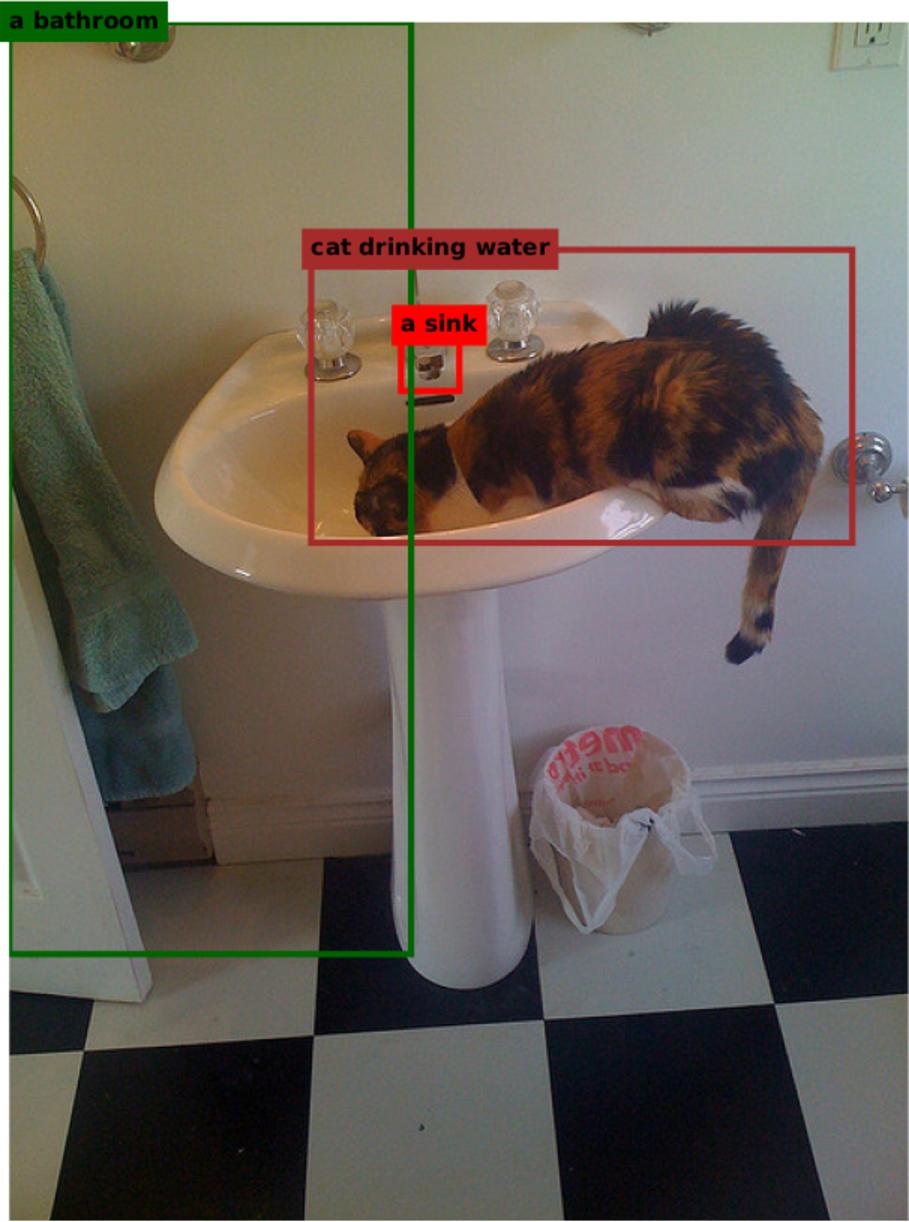}}
	    &
		\subfloat[\textcolor{BurntOrange}{golden  dog}  \textcolor{OliveGreen}{walking}  in  \textcolor{Fuchsia}{snow} with \textcolor{Maroon}{a person cross country} skiing the \textcolor{purple}{background}]{\includegraphics[width = 1.6in, height=1.25in]{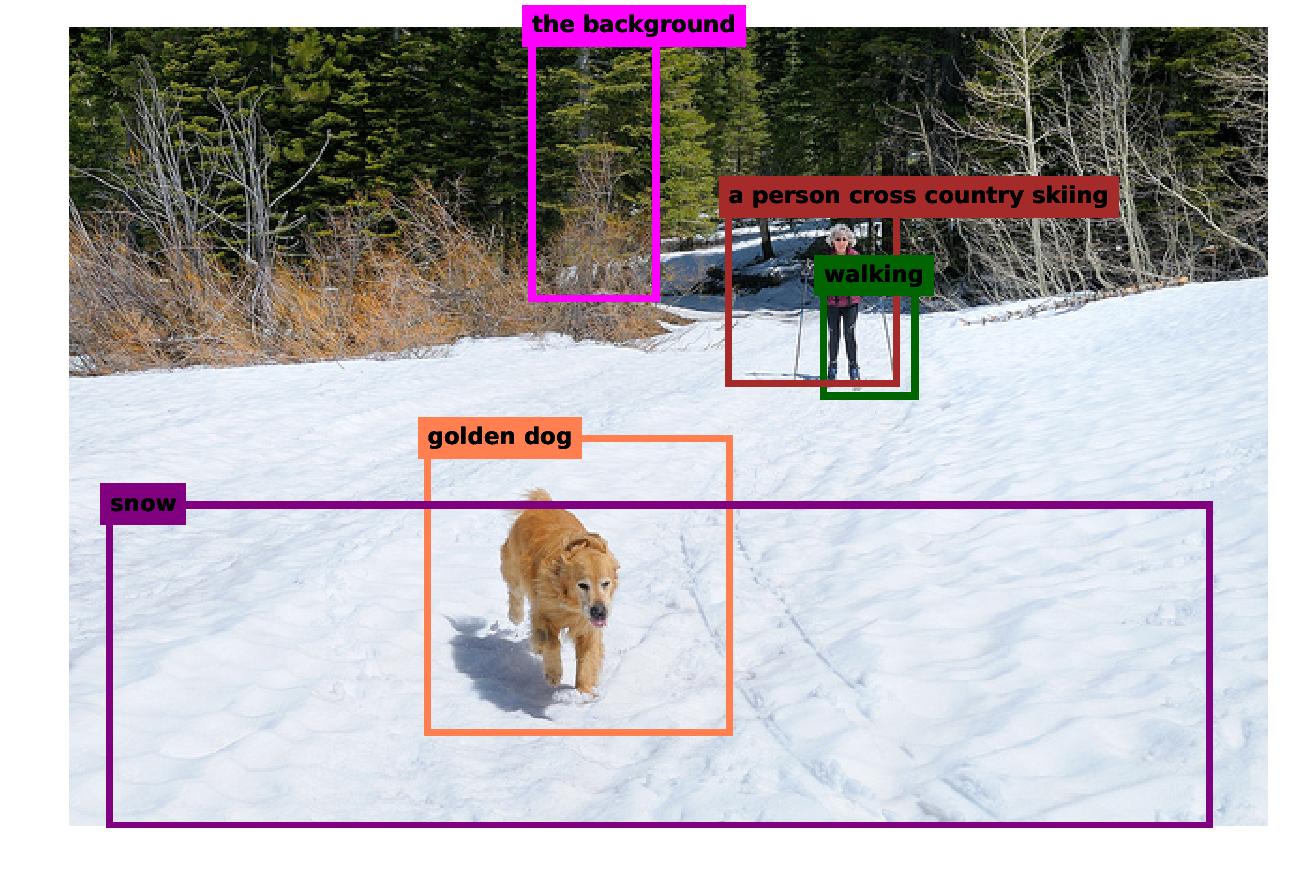}}
		&
    \subfloat[\small \textcolor{red}{a bath tub} \textcolor{GreenYellow}{sitting} next to \textcolor{cyan}{a sink} in \textcolor{OliveGreen}{a bathroom}]{\includegraphics[width = 1.6in, height=1.25in]{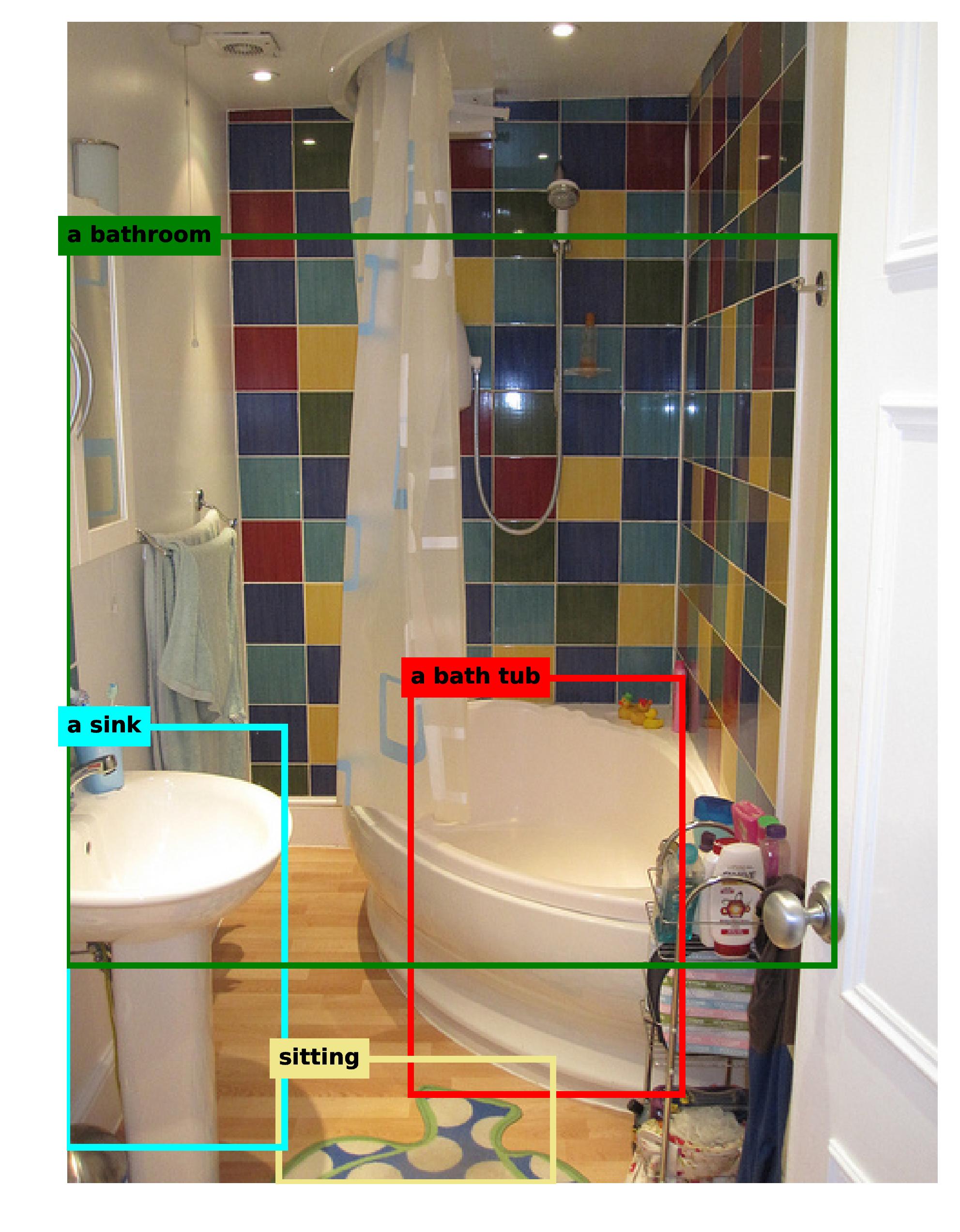}}
		&
		\subfloat[\small a person with \textcolor{OliveGreen}{a purple shirt} \textcolor{orange}{is painting} an image of \textcolor{blue}{a woman} on \textcolor{cyan}{a white wall}]{\includegraphics[width = 1.6in, height=1.25in]{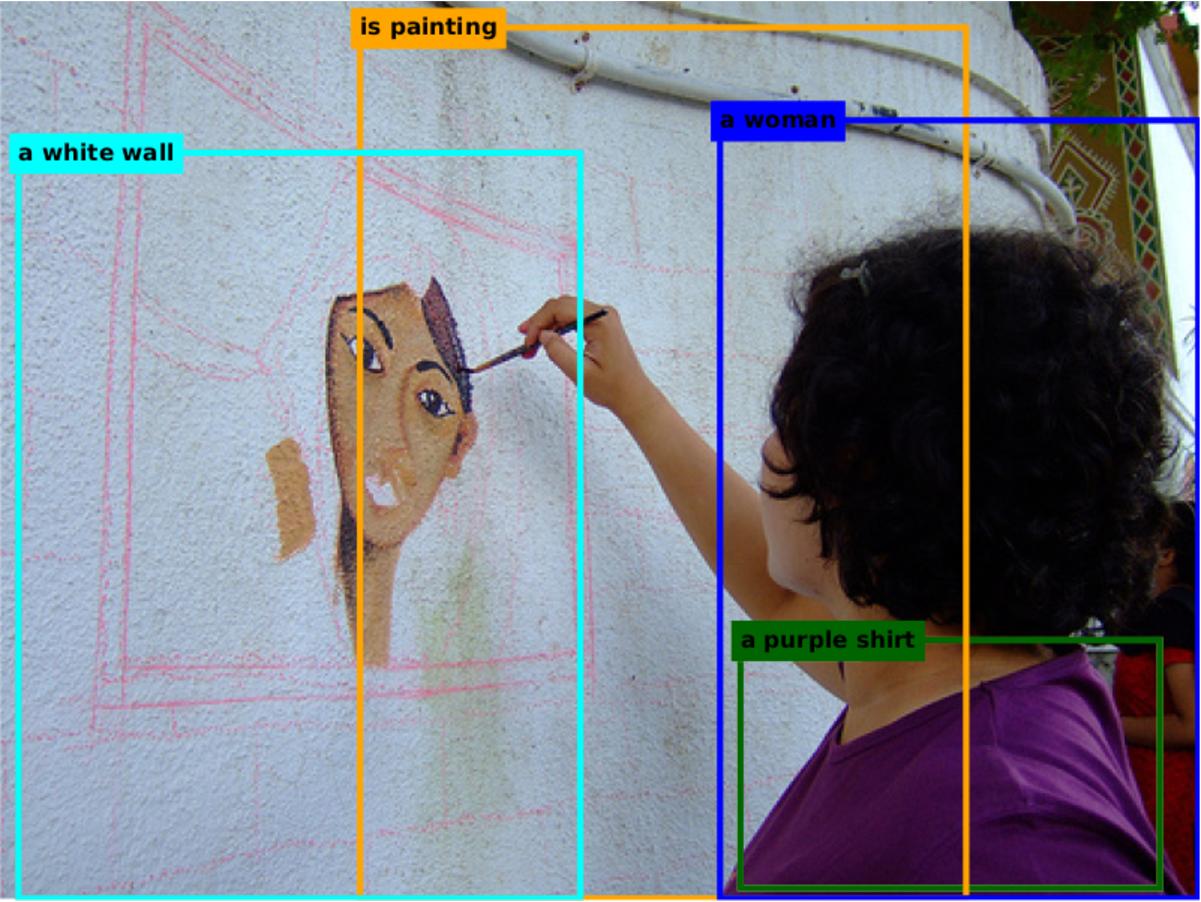}}
		\label {fig:detections}
	\end{tabular}
	\caption{\small We show the image regions that are matched with the phrases in the query caption for five image-caption pairs. Our model is able to effectively learn these correspondences without any phrase level ground-truth annotations during training. Figure best viewed in color.}
  \label{fig:detections}
  \vspace{-1.5em}
\end{figure*}

\begin{figure}[t]
  \centering
  \includegraphics[width=\columnwidth]{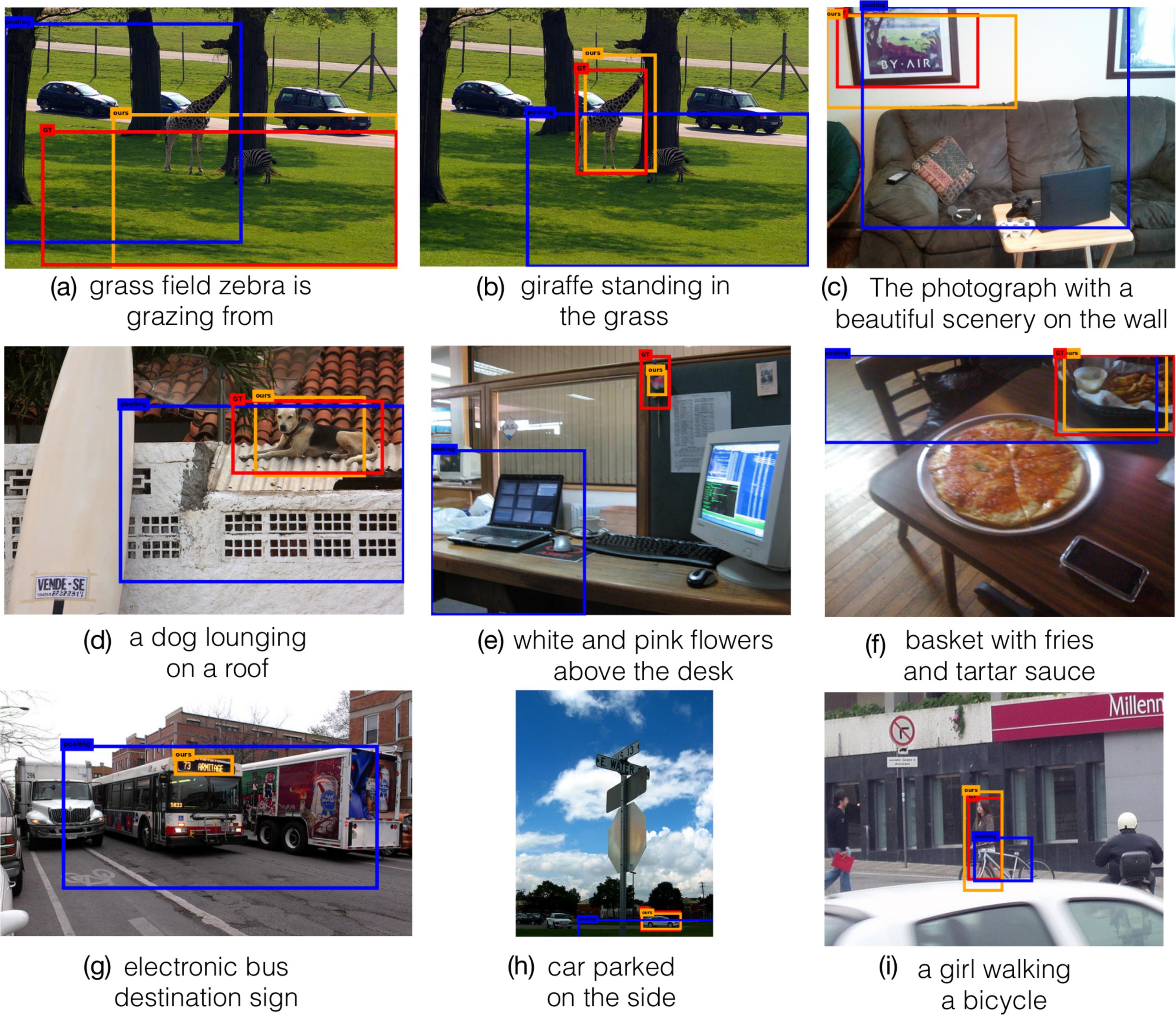}
  \caption{\small{We show outputs of our method Align2Ground (in orange) and Pooling-based method (in
	blue) on the phrase localization task for a few test images. The ground-truth is shown in red.
	Align2Ground is able to clearly localize better than
	the Pooling-based model in grounding noun only phrases \eg (e), (f) as well as phrases with
	verbs \eg (d), (i). Figure best viewed in color.}}
  \label{fig:local_phrases}
  \vspace{-1.5em}
\end{figure}

We observe that the Pooling-based (phrases) model ($R@1 = 48.4$) performs better on the C2I task than the Global baseline ($R@1 =
39.3$) on COCO (with the difference being even higher for Flickr30k). We also note that the Pooling-based (phrases)
outperforms its counterpart-- Pooling-based (words) that uses words for matching. This shows that for the C2I task, 
it is better to represent a caption as phrases instead of individual words, as used in this work. We also
notice improvements with the use of phrases on the phrase localization task ($Det\%$ $+0.1$ for COCO and $+0.5$
Flickr30k)

An interesting observation is that although the Pooling-based (phrases) method outperforms the Global baseline on the C2I
task, its performance on phrase localization is not always better than the latter ($Det\%$ $10.8$ vs.\ $12.2$ for
COCO and $8.9$ vs.\ $8.0$ for Flickr30k).  As stated in Section \ref{subsec-intro}, this trend could be explained by the
fact that on account of averaging the local matching scores, the Pooling-based methods are able to achieve good results
by selectively amplifying correspondences between phrases and image regions (e.g. by assigning high matching scores to
visual noun-phrases e.g. ``person'') without learning to accurately ground all phrases
(and ignoring less visual phrases e.g. ``sitting'') in the caption. Recall that this was one of the
motivations that inspired the design of our proposed Align2Ground model.

The Align2Ground model outperforms Global and Pooling-based baselines on both the datasets.  Specifically, we see
an improvement on the phrase localization performance, where our model yields better results than both the Global (by $+2.5$
on COCO and $+3.3$ on Flickr30k) and the Pooling-based (phrases) ($+3.9$ on COCO and $+2.8$ on Flickr30k) method.  We
believe that the superior performance of our model is due to the fact that our architectural design primes the
supervised loss to be a stronger learning signal for the phrase grounding task as compared to the Pooling-based
methods.  We also observe improvements on the C2I task of $8.2$ and $17.3$ on COCO compared to
the Pooling-based (phrases) and the Global methods respectively.

From our ablation studies on Align2Ground, we notice that the performance of our model is significantly influenced by
the choice of our Local Matching module. The \textit{topk} scheme consistently outperforms \textit{max} and
\textit{attention} schemes for both the datasets.  For example, when using topk for matching phrases with regions (\ie
permInv-topk), we see an increase (w.r.t.  using permInv-max) of $16.3$ and $20.6$ on $R@1$ for COCO and Flickr30k
respectively. We also observe similar trends when using the sequence encoder for encoding the matched RoIs.  These
results support our intuition that the introduction of randomness in the RoI selection step adds diversity to the
model and prevents overfitting by prematurely selecting a specific subset of RoIs --
a key issue in MIL~\cite{li2015multiple,cinbis2017weakly}.

\subsection{Qualitative Results}
\label{subsec-qualitative}

\begin{table}[t]
  \caption{Phrase Localization on Flicr30K Entities}
  \label{tab:loc-sota-flickr}
  \vspace{-1.5em}
  \renewcommand{\arraystretch}{0.5}
  \centering
  \setlength\tabcolsep{2pt}
	\resizebox{0.85\columnwidth}{!}{%
    \begin{tabular}{c c c c c c c}
      \multicolumn{7}{c}{} \\
      \toprule
      Akbari et al.~\cite{akbari2019multi} && \multirow{2}{*}{Fang et al.~\cite{fang2015captions}} && Pooling-based && \multirow{2}{*}{\textbf{Ours}} \\
      (prior SoTA) && && (phrases) && \\
      \midrule
      {$69.7$} && $29.0$ && $65.7$ && $\textbf{71.0}$ \\
      \bottomrule
    \end{tabular}%
  }
  \vspace{-1.5em}
\end{table}

In Figure \ref{fig:detections}, we show qualitative results of visual grounding of phrases performed
by our learned Local Matching module on a few test image--caption pairs (from COCO and Flickr30k).
From Figure \ref{fig:detections}, it is evident that our model is able to localize noun phrases
(e.g. ``white wall'', ``large pizza'') as well as verb phrases (e.g. ``holding'', ``standing'') from the query captions.

In Figure \ref{fig:local_phrases}, we show qualitative examples of phrase localization from the VG dataset.
We compare results of our model with those from Pooling-based methods.
We observe that our model is able to correctly localize phrases even when they appear in the midst of visual clutter.
For example, in image (f), our model is able to ground the phrase ``a basket of fries with tartar sauce''.
Another interesting example is the grounding of the phrase ``white and pink flowers above the desk'' in (e) where the
Pooling-based method gets confused and grounds the desk instead of the flowers. However, our model is able to correctly localize
the main subject of the phrase.

\subsection{Comparison with state-of-the-art}

We compare Align2Ground with state-of-the-art methods from literature for both the tasks of phrase localization
(Table \ref{tab:loc-sota}, \ref{tab:loc-sota-flickr}) and caption-to-image retrieval (Table \ref{tab:c2i-sota}).  For phrase localization, we
outperform the previous state-of-the-art~\cite{engilberge2018finding}, which uses a variant of the Global method with a
novel spatial pooling step, by $4.9\%$ based on the $PointIt\%$ metric on VG.
On Flickr30k Entities, we out-perform prior state-of-the-art~\cite{akbari2019multi} with much
simpler encoders (ResNet+bi-GRU v/s PNASNet+ElMo).
For the caption-to-image retrieval task, we
also achieve state-of-the-art performance ($R@1$ of $49.7$ vs. $48.6$ by~\cite{lee2018stacked}) on Flickr30k dataset and
get competitive results relative to state-of-the-art ($R@1$ of $56.6$ vs.\ $58.8$ by~\cite{lee2018stacked}) on COCO
dataset for the downstream C2I task.  These performance gains demonstrate that our model not only effectively learns to
ground phrases from the downstream C2I task, but the the tight coupling between these two also ends up helping the
downstream task (C2I retrieval).
\vspace{\sectionReduceBot}

\section{Conclusion}
In this work, we have proposed a novel method for phrase grounding using weak supervision
available from matching image--caption pairs.  Our key
contribution lies in designing the Local Aggregator module that is responsible for a tight coupling
between phrase grounding and image-caption matching
via caption--conditioned image representations.
We show that such an interaction between the two tasks primes the loss to
provide a stronger supervisory signal for phrase localization.
We report improvements of $4.9\%$ and $1.3\%$ (absolute) for
phrase localization on VG and Flickr30k Entities.  We also show
improvements of $3.9\%$ and $2.8\%$ on COCO and Flickr30k respectively
compared to prior methods.
This highlights the strength of the proposed representation in not allowiing
the model to get away without learning to ground all phrases and also
not compromising on the downstream task performance. Qualitative visualizations
of phrase grounding shows that our
model is able to effectively localize free-form phrases in images.

\vspace{\sectionReduceBot}

\section*{Acknowledgements}
The SRI effort was supported in part by the USAMRAA under Contract No. W81XWH-17-C-0083. The Georgia Tech
effort was supported in part by NSF, AFRL, DARPA, ONR YIPs.
The views, opinions, findings and/or conclusions contained herein are those of the authors and
should not be interpreted as necessarily representing the official policies or endorsements, either
expressed or implied, of the U.S. Government, or any sponsor.

{\small
\bibliographystyle{ieee_fullname}
\bibliography{biblio}

\begin{thebibliography}{10}\itemsep=-1pt

\bibitem{ahuja2018understanding}
Karuna Ahuja, Karan Sikka, Anirban Roy, and Ajay Divakaran.
\newblock Understanding visual ads by aligning symbols and objects using
  co-attention.
\newblock {\em arXiv preprint arXiv:1807.01448}, 2018.

\bibitem{akbari2019multi}
Hassan Akbari, Svebor Karaman, Surabhi Bhargava, Brian Chen, Carl Vondrick, and
  Shih-Fu Chang.
\newblock Multi-level multimodal common semantic space for image-phrase
  grounding.
\newblock {\em Conference on Computer Vision and Pattern Recognition}, 2019.

\bibitem{anderson2018bottom}
Peter Anderson, Xiaodong He, Chris Buehler, Damien Teney, Mark Johnson, Stephen
  Gould, and Lei Zhang.
\newblock Bottom-up and top-down attention for image captioning and visual
  question answering.
\newblock In {\em Conference on Computer Vision and Pattern Recognition},
  page~6, 2018.

\bibitem{bansal2018zero}
Ankan Bansal, Karan Sikka, Gaurav Sharma, Rama Chellappa, and Ajay Divakaran.
\newblock Zero-shot object detection.
\newblock In {\em European Conference on Computer Vision (ECCV)}, pages
  384--400, 2018.

\bibitem{bilen2014weakly}
Hakan Bilen, Marco Pedersoli, and Tinne Tuytelaars.
\newblock Weakly supervised object detection with posterior regularization.
\newblock In {\em British Machine Vision Conference}, volume~3, 2014.

\bibitem{chen2018knowledge}
Kan Chen, Jiyang Gao, and Ram Nevatia.
\newblock Knowledge aided consistency for weakly supervised phrase grounding.
\newblock {\em arXiv preprint arXiv:1803.03879}, 2018.

\bibitem{chung2015gated}
Junyoung Chung, Caglar Gulcehre, Kyunghyun Cho, and Yoshua Bengio.
\newblock Gated feedback recurrent neural networks.
\newblock In {\em International Conference on Machine Learning}, pages
  2067--2075, 2015.

\bibitem{cinbis2017weakly}
Ramazan~Gokberk Cinbis, Jakob Verbeek, and Cordelia Schmid.
\newblock Weakly supervised object localization with multi-fold multiple
  instance learning.
\newblock {\em Transactions on Pattern Analysis and Machine Intelligence},
  39(1):189--203, 2017.

\bibitem{cirik2018using}
Volkan Cirik, Taylor Berg-Kirkpatrick, and Louis-Philippe Morency.
\newblock Using syntax to ground referring expressions in natural images.
\newblock {\em arXiv preprint arXiv:1805.10547}, 2018.

\bibitem{collobert2011natural}
Ronan Collobert, Jason Weston, L{\'e}on Bottou, Michael Karlen, Koray
  Kavukcuoglu, and Pavel Kuksa.
\newblock Natural language processing (almost) from scratch.
\newblock {\em Journal of Machine Learning Research}, 12(Aug):2493--2537, 2011.

\bibitem{doersch2015unsupervised}
Carl Doersch, Abhinav Gupta, and Alexei~A Efros.
\newblock Unsupervised visual representation learning by context prediction.
\newblock In {\em International Conference on Computer Vision}, pages
  1422--1430, 2015.

\bibitem{eisenschtat2017linking}
Aviv Eisenschtat and Lior Wolf.
\newblock Linking image and text with 2-way nets.
\newblock In {\em Conference on Computer Vision and Pattern Recognition}, 2017.

\bibitem{engilberge2018finding}
Martin Engilberge, Louis Chevallier, Patrick P{\'e}rez, and Matthieu Cord.
\newblock Finding beans in burgers: Deep semantic-visual embedding with
  localization.
\newblock In {\em Conference on Computer Vision and Pattern Recognition}, pages
  3984--3993, 2018.

\bibitem{faghri2017vse++}
Fartash Faghri, David~J Fleet, Jamie~Ryan Kiros, and Sanja Fidler.
\newblock Vse++: improved visual-semantic embeddings.
\newblock {\em arXiv preprint arXiv:1707.05612}, 2017.

\bibitem{fang2015captions}
Hao Fang, Saurabh Gupta, Forrest Iandola, Rupesh~K Srivastava, Li Deng, Piotr
  Doll{\'a}r, Jianfeng Gao, Xiaodong He, Margaret Mitchell, John~C Platt,
  et~al.
\newblock From captions to visual concepts and back.
\newblock In {\em Proceedings of the IEEE conference on computer vision and
  pattern recognition}, pages 1473--1482, 2015.

\bibitem{frome2013devise}
Andrea Frome, Greg~S Corrado, Jon Shlens, Samy Bengio, Jeff Dean, Tomas
  Mikolov, et~al.
\newblock Devise: A deep visual-semantic embedding model.
\newblock In {\em Neural Information Processing Systems}, pages 2121--2129,
  2013.

\bibitem{he2017mask}
Kaiming He, Georgia Gkioxari, Piotr Doll{\'a}r, and Ross Girshick.
\newblock Mask r-cnn.
\newblock In {\em International Conference on Computer Vision}, pages
  2980--2988. IEEE, 2017.

\bibitem{he2016identity}
Kaiming He, Xiangyu Zhang, Shaoqing Ren, and Jian Sun.
\newblock Identity mappings in deep residual networks.
\newblock In {\em European Conference on Computer Vision}, pages 630--645.
  Springer, 2016.

\bibitem{huang2017instance}
Yan Huang, Wei Wang, and Liang Wang.
\newblock Instance-aware image and sentence matching with selective multimodal
  lstm.
\newblock In {\em Conference on Computer Vision and Pattern Recognition},
  page~7, 2017.

\bibitem{karpathy2015deep}
Andrej Karpathy and Li Fei-Fei.
\newblock Deep visual-semantic alignments for generating image descriptions.
\newblock In {\em Conference on Computer Vision and Pattern Recognition}, pages
  3128--3137, 2015.

\bibitem{karpathy2014deep}
Andrej Karpathy, Armand Joulin, and Li~F Fei-Fei.
\newblock Deep fragment embeddings for bidirectional image sentence mapping.
\newblock In {\em Neural Information Processing Systems}, pages 1889--1897,
  2014.

\bibitem{kiros2014unifying}
Ryan Kiros, Ruslan Salakhutdinov, and Richard~S Zemel.
\newblock Unifying visual-semantic embeddings with multimodal neural language
  models.
\newblock {\em arXiv preprint arXiv:1411.2539}, 2014.

\bibitem{kottur2017natural}
Satwik Kottur, Jos{\'e}~MF Moura, Stefan Lee, and Dhruv Batra.
\newblock Natural language does not emerge'naturally'in multi-agent dialog.
\newblock {\em arXiv preprint arXiv:1706.08502}, 2017.

\bibitem{krishna2017visual}
Ranjay Krishna, Yuke Zhu, Oliver Groth, Justin Johnson, Kenji Hata, Joshua
  Kravitz, Stephanie Chen, Yannis Kalantidis, Li-Jia Li, David~A Shamma, et~al.
\newblock Visual genome: Connecting language and vision using crowdsourced
  dense image annotations.
\newblock {\em International Journal of Computer Vision}, 123(1):32--73, 2017.

\bibitem{lee2018stacked}
Kuang-Huei Lee, Xi Chen, Gang Hua, Houdong Hu, and Xiaodong He.
\newblock Stacked cross attention for image-text matching.
\newblock {\em arXiv preprint arXiv:1803.08024}, 2018.

\bibitem{li2015multiple}
Weixin Li and Nuno Vasconcelos.
\newblock Multiple instance learning for soft bags via top instances.
\newblock In {\em Conference on Computer Vision and Pattern Recognition}, pages
  4277--4285, 2015.

\bibitem{lin2014microsoft}
Tsung-Yi Lin, Michael Maire, Serge Belongie, James Hays, Pietro Perona, Deva
  Ramanan, Piotr Doll{\'a}r, and C~Lawrence Zitnick.
\newblock Microsoft coco: Common objects in context.
\newblock In {\em European Conference on Computer Vision}, pages 740--755.
  Springer, 2014.

\bibitem{ma2015multimodal}
Lin Ma, Zhengdong Lu, Lifeng Shang, and Hang Li.
\newblock Multimodal convolutional neural networks for matching image and
  sentence.
\newblock In {\em International Conference on Computer Vision}, pages
  2623--2631, 2015.

\bibitem{mao2016generation}
Junhua Mao, Jonathan Huang, Alexander Toshev, Oana Camburu, Alan~L Yuille, and
  Kevin Murphy.
\newblock Generation and comprehension of unambiguous object descriptions.
\newblock In {\em Conference on Computer Vision and Pattern Recognition}, pages
  11--20, 2016.

\bibitem{mao2014deep}
Junhua Mao, Wei Xu, Yi Yang, Jiang Wang, Zhiheng Huang, and Alan Yuille.
\newblock Deep captioning with multimodal recurrent neural networks (m-rnn).
\newblock {\em arXiv preprint arXiv:1412.6632}, 2014.

\bibitem{misra2016shuffle}
Ishan Misra, C~Lawrence Zitnick, and Martial Hebert.
\newblock Shuffle and learn: unsupervised learning using temporal order
  verification.
\newblock In {\em European Conference on Computer Vision}, pages 527--544.
  Springer, 2016.

\bibitem{nam2016dual}
Hyeonseob Nam, Jung-Woo Ha, and Jeonghee Kim.
\newblock Dual attention networks for multimodal reasoning and matching.
\newblock {\em arXiv preprint arXiv:1611.00471}, 2016.

\bibitem{niu2017hierarchical}
Zhenxing Niu, Mo Zhou, Le Wang, Xinbo Gao, and Gang Hua.
\newblock Hierarchical multimodal lstm for dense visual-semantic embedding.
\newblock In {\em International Conference on Computer Vision}, pages
  1899--1907. IEEE, 2017.

\bibitem{plummer2015flickr30k}
Bryan~A Plummer, Liwei Wang, Chris~M Cervantes, Juan~C Caicedo, Julia
  Hockenmaier, and Svetlana Lazebnik.
\newblock Flickr30k entities: Collecting region-to-phrase correspondences for
  richer image-to-sentence models.
\newblock In {\em International Conference on Computer Vision}, pages
  2641--2649, 2015.

\bibitem{qi2017pointnet}
Charles~R Qi, Hao Su, Kaichun Mo, and Leonidas~J Guibas.
\newblock Pointnet: Deep learning on point sets for 3d classification and
  segmentation.
\newblock In {\em Conference on Computer Vision and Pattern Recognition}, pages
  652--660, 2017.

\bibitem{ren2015faster}
Shaoqing Ren, Kaiming He, Ross Girshick, and Jian Sun.
\newblock Faster r-cnn: Towards real-time object detection with region proposal
  networks.
\newblock In {\em Neural Information Processing Systems}, pages 91--99, 2015.

\bibitem{rohrbach2016grounding}
Anna Rohrbach, Marcus Rohrbach, Ronghang Hu, Trevor Darrell, and Bernt Schiele.
\newblock Grounding of textual phrases in images by reconstruction.
\newblock In {\em European Conference on Computer Vision}, pages 817--834.
  Springer, 2016.

\bibitem{vendrov2015order}
Ivan Vendrov, Ryan Kiros, Sanja Fidler, and Raquel Urtasun.
\newblock Order-embeddings of images and language.
\newblock {\em arXiv preprint arXiv:1511.06361}, 2015.

\bibitem{wang2018learning}
Liwei Wang, Yin Li, Jing Huang, and Svetlana Lazebnik.
\newblock Learning two-branch neural networks for image-text matching tasks.
\newblock {\em Transactions on Pattern Analysis and Machine Intelligence},
  2018.

\bibitem{xiao2017weakly}
Fanyi Xiao, Leonid Sigal, and Yong~Jae Lee.
\newblock Weakly-supervised visual grounding of phrases with linguistic
  structures.
\newblock {\em arXiv preprint arXiv:1705.01371}, 2017.

\bibitem{ye2007adaptive}
Jieping Ye, Zheng Zhao, and Huan Liu.
\newblock Adaptive distance metric learning for clustering.
\newblock In {\em Conference on Computer Vision and Pattern Recognition}, pages
  1--7. IEEE, 2007.

\bibitem{young2014image}
Peter Young, Alice Lai, Micah Hodosh, and Julia Hockenmaier.
\newblock From image descriptions to visual denotations: New similarity metrics
  for semantic inference over event descriptions.
\newblock {\em Transactions of the Association for Computational Linguistics},
  2:67--78, 2014.

\bibitem{zaheer2017deep}
Manzil Zaheer, Satwik Kottur, Siamak Ravanbakhsh, Barnabas Poczos, Ruslan~R
  Salakhutdinov, and Alexander~J Smola.
\newblock Deep sets.
\newblock In {\em Neural Information Processing Systems}, pages 3391--3401,
  2017.

\bibitem{zhang2018top}
Jianming Zhang, Sarah~Adel Bargal, Zhe Lin, Jonathan Brandt, Xiaohui Shen, and
  Stan Sclaroff.
\newblock Top-down neural attention by excitation backprop.
\newblock {\em International Journal of Computer Vision}, 126(10):1084--1102,
  2018.

\bibitem{zhu2016visual7w}
Yuke Zhu, Oliver Groth, Michael Bernstein, and Li Fei-Fei.
\newblock Visual7w: Grounded question answering in images.
\newblock In {\em Conference on Computer Vision and Pattern Recognition}, pages
  4995--5004, 2016.

\end{thebibliography}
}

\end{document}